\def\year{2020}\relax
\documentclass[letterpaper]{article} % DO NOT CHANGE THIS
\usepackage{aaai20}  % DO NOT CHANGE THIS
\usepackage{times}  % DO NOT CHANGE THIS
\usepackage{helvet} % DO NOT CHANGE THIS
\usepackage{courier}  % DO NOT CHANGE THIS
\usepackage[hyphens]{url}  % DO NOT CHANGE THIS
\usepackage{graphicx} % DO NOT CHANGE THIS
\usepackage{graphicx}  % DO NOT CHANGE THIS
\usepackage{amsmath,amssymb} % define this before the line numbering.
\usepackage{multirow}
\usepackage{subfigure}
\usepackage{array}
\usepackage{xcolor}
\usepackage{diagbox}

\def\bm #1{\boldsymbol{#1}}%%%%%matrix
\makeatletter
\DeclareRobustCommand\onedot{\futurelet\@let@token\@onedot}
\def\@onedot{.}
\def\eg{{e.g}\onedot} 
\def\ie{{i.e}\onedot}

\def\etal{\textit{et al}\onedot~}
\makeatother
\newcommand{\Tref}[1]{Table~\ref{#1}}

\newcommand{\Fref}[1]{Figure~\ref{#1}}

\newcommand{\fref}[1]{Fig.~\ref{#1}}

\title{A Coarse-to-Fine Adaptive Network for Appearance-Based Gaze Estimation}
\author{Yihua Cheng\textsuperscript{\rm 1}, Shiyao Huang\textsuperscript{\rm 2}, Fei Wang\textsuperscript{\rm 2}, Chen Qian\textsuperscript{\rm 2}, Feng Lu\textsuperscript{\rm 1,3,4} % All authors must be in the same font size and format. Use \Large and \textbf to achieve this result when breaking a line
	\thanks{Feng Lu is the Corresponding Author.\newline This work was supported by National Natural Science Foundation of China (NSFC) under Grant 61972012.}\\
	\textsuperscript{\rm 1}State Key Laboratory of Virtual Reality Technology and Systems, SCSE, Beihang University, China.\\
	\textsuperscript{\rm 2}SenseTime Co., Ltd.\quad \quad\quad\quad\quad\quad~\textsuperscript{\rm 3}Peng Cheng Laboratory, Shenzhen, China.\\
	\textsuperscript{\rm 4}Beijing Advanced Innovation Center for Big Data-Based Precision Medicine,
	Beihang University,  China.\\
	\{yihua\_c, lufeng\}@buaa.edu.cn, \{huangshiyao, wangfei, qianchen\}@sensetime.com  % email address must be in roman text type, not monospace or sans serif
}
 
\begin{document}

\maketitle

\begin{abstract}
	Human gaze is essential for various appealing applications. 
	Aiming at more accurate gaze estimation, a series of recent works propose to utilize face and eye images simultaneously. 
	Nevertheless, face and eye images only serve as independent or parallel feature sources in those works, the intrinsic correlation between their features is overlooked. In this paper we make the following contributions: 1) We propose a coarse-to-fine strategy which estimates a basic gaze direction from face image and refines it with corresponding residual predicted from eye images. 2) Guided by the proposed strategy, we design a framework which introduces a bi-gram model to bridge gaze residual and basic gaze direction, and an attention component to adaptively acquire suitable fine-grained feature. 3) Integrating the above innovations, we construct a coarse-to-fine adaptive network named CA-Net and achieve state-of-the-art performances on MPIIGaze and EyeDiap.
\end{abstract}

\section{Introduction}
Human gaze implicates important cues for applications such as 
saliency detection~\cite{2018_alshawi_saliency}, human-computer interaction~\cite{2017_zhang_contact} and virtual reality~\cite{2016_anjul_vr}.  

Gaze estimation methods can be divided into model-based methods and appearance-based methods. Model-based methods generally achieve accurate gaze estimation with dedicated devices, but are mostly limited to laboratory environment due to short working distance (typically within 60cm) and high failure rate  in the wild. Appearance-based methods attract much attention recently, they require only a webcam to capture images and directly learn the mapping from images to gaze directions.
As human eye appearance can be influenced by various factors in the wild such as head pose, CNN-based methods are proposed and significantly outperform classical methods thanks to CNN's superior ability in learning very complex mapping functions. 

\begin{figure}[t]
	\begin{center}
		\includegraphics[width=\columnwidth]{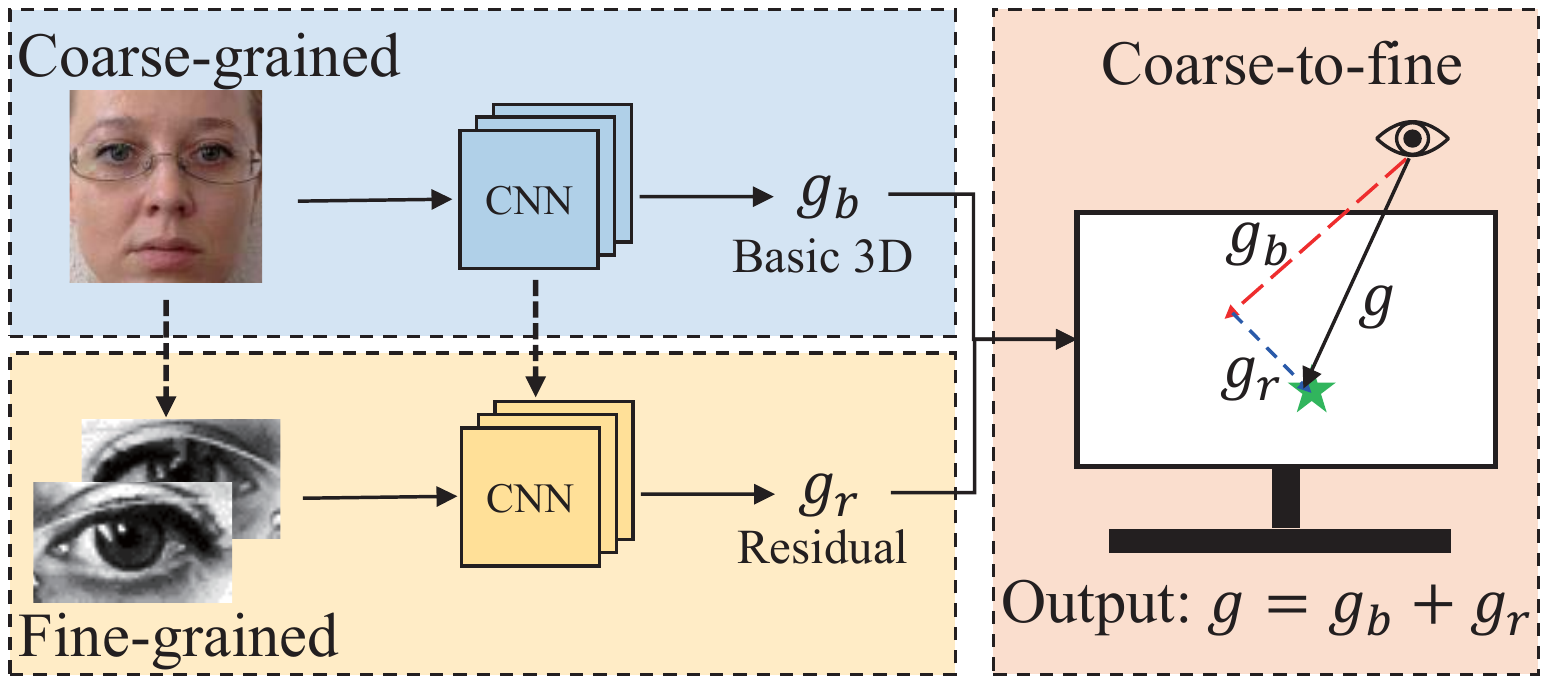}	
	\end{center}
	\caption{The process of coarse-to-fine strategy. We extract coarse-grained feature from face image to estimate basic gaze direction $g_b$ and extract fine-grained feature from eye images to estimate gaze residual $g_r$. We use $g_r$ to refine $g_b$ and acquire the outputted gaze direction $g$.}
	\label{fig:overview}
\end{figure}  

CNN-based methods estimate gaze directions from face or eye images.
Zhang \etal~\cite{2015_zhang_appearancewild} first propose a network to estimate gaze directions from eye images. Afterwards face images are put forward~\cite{2017_zhang_fullface}.
Recently, methods simultaneously utilizing face and eye images achieve even better results~\cite{2018_zhao_dilated}. Nevertheless, previous methods treat face and eye images only as independent or parallel feature sources, thus neglecting their intrinsic relationship at the level of feature granularity. In fact, eye image provide fine-grained feature focusing on gaze, while face image supplies coarse-grained feature with richer information.

To make full use of the feature relationship between face and eyes, we propose a coarse-to-fine strategy in this paper, which achieve state-of-the-art performances on the most common benchmarks. The core idea of proposed coarse-to-fine strategy is to \textbf{estimate a basic gaze direction from face image and refine it with corresponding residual predicted from eye images.}
Specifically, since face image carries richer information than eye images, we utilize it to estimate an approximation of gaze direction. Then we extract fine-grained feature from eye images to refine the basic gaze direction and finally acquire coarse-to-fine gaze direction.

As shown in~\Fref{fig:overview}, we first design a CNN to extract coarse-grained feature from face image and predict the basic gaze direction. Next, another CNN is set up to extract the fine-grained feature from eye images and generate gaze residual. At last, we acquire the final gaze direction by adding the basic gaze direction and gaze residual vectorially.

However, there still remain two key problems to be considered. 
The first problem is how to ensure the estimated gaze residual is effective for refining its corresponding base gaze direction.
The second problem is how to enforce the fine-grained features extracted from eye images to be suitable for estimating the gaze residual.
Inspired from NLP algorithms, we generalize the coarse-to-fine process as a bi-gram model to solve the first problem. The bi-gram model bridges gaze residual and basic gaze direction, and produces gaze residual coupling with basic gaze direction. 
For the second problem, an attention component is proposed to adaptively acquire suitable fine-grained features.

Integrating above algorithms, we finally propose the coarse-to-fine adaptive network (CA-Net) for gaze estimation, which can adaptively acquire suitable fine-grained feature and estimates 3D gaze directions in a coarse-to-fine way.
To the best of our knowledge, we are the first to consider the intrinsic correlation between face and eye images and propose a framework for coarse-to-fine gaze estimation.

In summary, the contributions of this work are threefold:
\begin{enumerate}
	\item We propose a novel coarse-to-fine strategy for gaze estimation. 
	\item We propose an ingenious framework for coarse-to-fine gaze estimation. The framework introduces a bi-gram model, which bridges coarse-grained gaze estimation and fine-grained gaze estimation, and an attention component, which adaptively acquires suitable fine-grained feature.
	\item Based on the proposed framework, we design a network named CA-Net and achieve state-of-the-art performances on MPIIGaze and EyeDiap.
\end{enumerate}

\section{Related work}
Gaze estimation methods can be simply divided into model-based and appearance-based~\cite{2010_hansen_survey}.

\subsection{Model-based methods}
Model-based methods can estimate gaze with good accuracy by building geometric eye models~\cite{2006_guestrin_remote}.
They typically fit the model by detecting eye features such as near infrared corneal reflections~\cite{2012_nakazawa_point}, pupil center~\cite{2012_valenti_combineheadpose}, and iris contours~\cite{2014_mora_geometric,2014_xiong_rgbdcamera}.
However, the detection of eye features may require dedicated devices such as infrared lights, stereo/high-definition cameras, and RBG-D cameras~\cite{2014_mora_geometric,2014_xiong_rgbdcamera}. 
Meanwhile, model-based methods usually have limited working distances between the user and the camera. 
These limitations show that model-based methods are more suitable for controlled environment, \eg, the laboratory setting, rather than outdoor settings.

\subsection{Appearance-based methods}
Most of appearance-based methods only require a webcam to capture images and learn the mapping function from images to the corresponding gaze~\cite{2002_karhan_appearancebased}.
The loose requirement attracts much attention for appearance-based methods.
Up to now, many methods such as Neural networks ~\cite{1994_baluja_firstneural,1998_xu_novel}, adaptive linear regression~\cite{2014_lu_adaptive}, Gaussian process regression~\cite{2006_williams_sparse} and dimension reduction~\cite{2017_lu_uncalibrated} have been proposed to learn the mapping function.
In order to handle arbitrary head motion, images can be used to learn more complex mapping functions~\cite{2014_lu_headmotion,2015_lu_synthesis}.
However, learning a generic mapping function is still challenging because of the highly non-liner of mapping function.

Recently, CNNs-based methods show better accuracy than conventional appearance methods. 
Zhang~\etal~\cite{2015_zhang_appearancewild} first proposed a CNNs-based method to estimate gaze, the method was designed based on LeNet~\cite{1998_lecun_lenet} and estimates gaze from eye images.
Yu~\etal~\cite{2018_yu_deep} proposed a multitask gaze estimation model with landmark constrain, they estimate gaze from eye images.
Fischer~\etal~\cite{2018_tobias_rt} extracted feature from two-eye images with VGG-16~\cite{2014_karen_vgg} to estimate gaze, they use an ensemble scheme to increase robustness of proposed method.
Cheng~\etal~\cite{2018_cheng_asymmetric} proposed a CNNs-based network which uses two-eye images as inputs and utilizes the two-eye asymmetry to optimize whole network.

Meanwhile, recent studies prove face images is effective in CNNs-based methods.
Krafka~\etal~\cite{2016_krafka_itrack} implemented the CNNs-based gaze tracker in the mobile devices, it estimates gaze from face and eye images.
Zhang~\etal~\cite{2017_zhang_fullface} proposed a spatial weights CNN to estimate gaze from face images.  
Deng \etal~\cite{2017_zhu_monocular} proposed a CNNs-based method with geometry constraints, it uses face and eye images as inputs and can estimate gaze in free-head setting.
Zhao \etal~\cite{2018_zhao_dilated} proposed a CNNs-based method using dilated convolution to estimate gaze from face and eye images.
Xiong~\etal~\cite{2019_xiong_mixed} combines the mixed effects model with CNN and estimates gaze from face images.  

\section{Method}
In this section, we introduce the architecture of our CA-Net, 
which can adaptively acquire suitable fine-grained feature and estimate the gaze directions in a coarse-to-fine way.

\subsection{Overview}
The core idea of the coarse-to-fine strategy is to estimate a basic gaze direction from face image and refine it with corresponding residual predicted from eye images. We propose the CA-Net based on the coarse-to-fine strategy.

The CA-Net contains two subnets: Face-Net and Eye-Net.
Face-Net extracts coarse-grained feature from face image and estimates the basic gaze direction.  
Eye-Net estimates gaze residual from two eye images to refine the basic gaze direction.
Next, We first propose an attention component to adaptively assign weights for two-eye features. 
A suitable eye feature is acquired by adding the weighted two-eye features together.
In addition, since the gaze residual is associated with the basic gaze estimation, we generalize the coarse-to-fine process as a bi-gram model to bridge the Face-net and Eye-net, and produces gaze residual coupling with basic gaze direction.
Finally, CA-Net output gaze direction by adding the basic gaze direction and gaze residual together.

The rests of this section are organized as follows. 
We first introduce the process of feature generation, in which we propose an attention component to adaptively assign weights for two-eye features.
We then introduce the coarse-to-fine strategy, which can be generalized as a bi-gram model.
Next, we detail the architecture of proposed CA-Net and define the loss function of CA-Net. 
At last, we present the implementation details at the end of this section.
\begin{figure}[t]
	\begin{center}
		\includegraphics[width=\columnwidth]{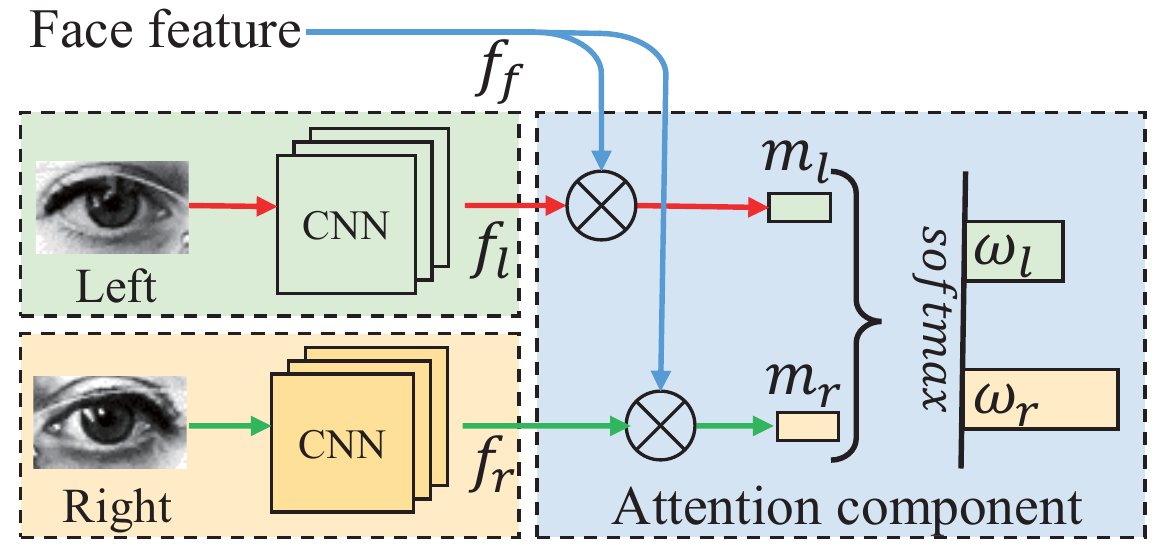}	
	\end{center}
	\caption{The architecture of proposed attention component. It adaptively assigns weights for left and right eyes.}
	\label{fig:attention}
\end{figure}

\subsection{Feature generation}
A key point of coarse-to-fine strategy is to acquire suitable feature, especially for estimating gaze residual. Therefore, we first describe the process of feature generation.

The face feature is used to estimate basic gaze direction.
Therefore, a common CNN is used to extract the coarse-grained face feature from face images. 
As for eye feature, we also respectively use a CNN to extract features from two eye images.
However, after acquiring the left and right eye features, a key problem is how to obtain suitable eye feature from two-eye features to accurately estimate gaze residual. 

There are at least two factors we need to consider.
First, as for different basic gaze directions, the suitable eye features can be different.
Second, two eye appearances have different reliabilities for gaze estimation~\cite{2018_cheng_asymmetric} because of the in-the-wild setting such as free-head. 
Those two factors both influence the acquirement of suitable eye feature.
In order to tackle above factors, we propose an attention component which can adaptively assign weights for two eyes.
The suitable eye feature is produced by summing the weighted left and right eye features. 

The attention component is inspired by attention mechanisms, which are widely used in NLP~\cite{2017_ash_attention}. The architecture of proposed component is shown in~\fref{fig:attention}.
In particular, as for left eye image, a score $m_l$ is produced by left eye feature $f_{l}$ and face feature $f_{f}$. 
As for right eye image, a score $m_r$ is produced by right eye feature $f_{r}$ and face feature $f_{f}$.
Then, a softmax layer is used to balance the scores $m_l$ and $m_r$ and the weights $w_l$ and $w_r$ are outputted for left eye and right eye.
The method which produces scores from feature is various, in the paper, we directly use the method proposed in~\cite{2015_bah_attention}.

The proposed component has following properties:
\begin{enumerate}
	\item Score $m_l$ is related with face feature, which is used to predict basic gaze directions. It means that basic gaze directions can decide the size of $m_l$. This corresponds with the first factor we describe above.
	\item Score $m_l$ is related with left eye feature. On the other words, $m_l$ is related with the left eye appearance. It corresponds with the second factor we describe above.
	\item Score $m_l$ is irrelevant to right eye feature, it is reasonable that the scores of left eyes are irrelevant to right eyes.
	\item $w_l$ is generated by comparing $m_l$ and $m_r$. Although the scores of left eyes are irrelevant to right eyes, the final weight should be generated with considering both the scores of two eyes.
	\item Above properties also suit for $m_r$ and $w_r$.
\end{enumerate}  

We also formulate the process of our implementation as follows.
we acquire the score of left eyes by
\begin{equation}
	\label{equ:weight_l}
	m_l = v^Ttanh(W_1^Tf_{f} + W_2^Tf_{l}),
\end{equation}
and acquire the score of right eyes by
\begin{equation}
	\label{equ:weight_r}
	m_r = v^Ttanh(W_1^Tf_{f} + W_2^Tf_{r}),
\end{equation}
where $v$, $W_1$ and $W_2$ are learned parameters and are implemented by fully connected layers.

Meanwhile, a softmax layer is used to balance the scores of left and right and outputs the weights.
\begin{equation}
	\label{equ:softmax}
	[w_l, w_r] = softmax([m_l, m_r]).
\end{equation}

The final eye feature $f_{e}$ can be acquired by
\begin{equation}
	\label{equ:eyefeature}
	f_{e} = w_l*f_{l} + w_r*f_{r}.
\end{equation}
\begin{figure}[t]
	\begin{center}
		
		\label{fig:accuracy-c}  
		\includegraphics[width=0.6\columnwidth]{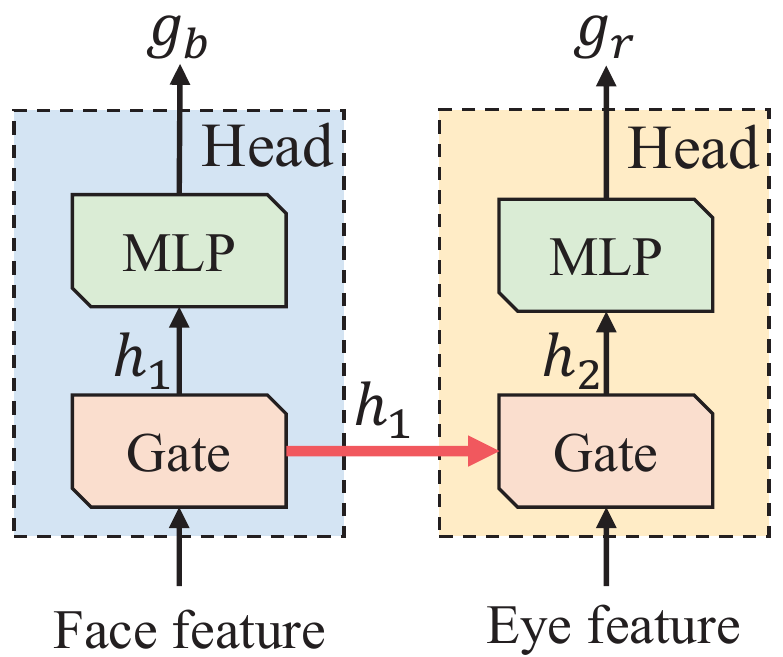}	
		
	\end{center}
	\caption{The coarse-to-fine process is generalized as a bi-gram model, which bridges gaze residual $g_r$ and corresponding basic gaze direction $g_b$. }
	\label{fig:coarse2fine}
\end{figure} 

\subsection{Coarse-to-fine gaze estimation}
% Therefore,
After acquiring the features, it is still unknown how to perform gaze estimation in a coarse-to-fine way.
A straightforward solution is that learn a mapping function to estimate the basic gaze direction from coarse-grained face feature and then learn another mapping function to estimate the gaze residual from fine-grained eye feature.
However, this solution has two problems. 
First, it do not consider the relation between basic gaze directions and gaze residuals. 
Second, for the estimation of gaze residual, although eye images are finer than face images, it is bad to directly discard the face feature.
Therefore, we generalize the coarse-to-fine process as a bi-gram model. 
The architecture of the bi-gram model is shown in \fref{fig:coarse2fine}, we omit the process of feature generation which can be various.

Specifically, as show in \fref{fig:coarse2fine}, the face feature is processed by a gate function to produce state $h_1$. Then, on one hand, the state $h_1$ is used to estimate the basic gaze direction. On the other hand, the state $h_1$ is delivered into the next gate and produces the state $h_2$ with eye feature. The state $h_2$ is used to estimate gaze residuals.
The gate function can be various. The main task of designed gate is to filter the previous states and reduce the influence of previous task on current task.   
We use GRU~\cite{2014_cho_gru} in this work.

The coarse-to-fine process can be understood as follows.
The basic gaze directions are directly estimated from state $h_1$, which is produced by face feature.
The gaze residuals are estimated from state $h_2$, which is generated from $h_1$ and eye feature.
This means the process of estimating gaze residuals is related with basic gaze directions.
Meanwhile, with delivering $h_1$, the face feature is also implicitly used to estimate gaze residuals rather than discarding.
Moreover, since the face feature includes much coarse-grained information, a learned gate is used to adaptively filter the state $h_1$. 

The process of learned gate is shown as follow: 
\begin{equation}
	\label{equ:forget}
	z_i = \sigma(W_z \cdot [h_i, f]).
\end{equation}

\begin{equation}
	\label{equ:reset}
	r_i = \sigma(W_r \cdot [h_i, f]).
\end{equation}

\begin{equation}
	\label{equ:newfeature}
	\tilde{h_{i+1}} = ReLU(W_h \cdot [r_i \ast h_i, f]).
\end{equation}

\begin{equation}
	\label{equ:generation}
	h_{i+1} = (1-z_i) \ast h_i + z_i \ast \tilde{h_{i+1}},
\end{equation}

where $f$ represents the corresponding feature. $W_z$, $W_r$ and $W_h$ are learned parameters, which can be implemented with fully connected layers. The $h_0$ is set as a zero matrix.
\begin{figure}[t]
	\begin{center}
		\includegraphics[width=0.9\columnwidth]{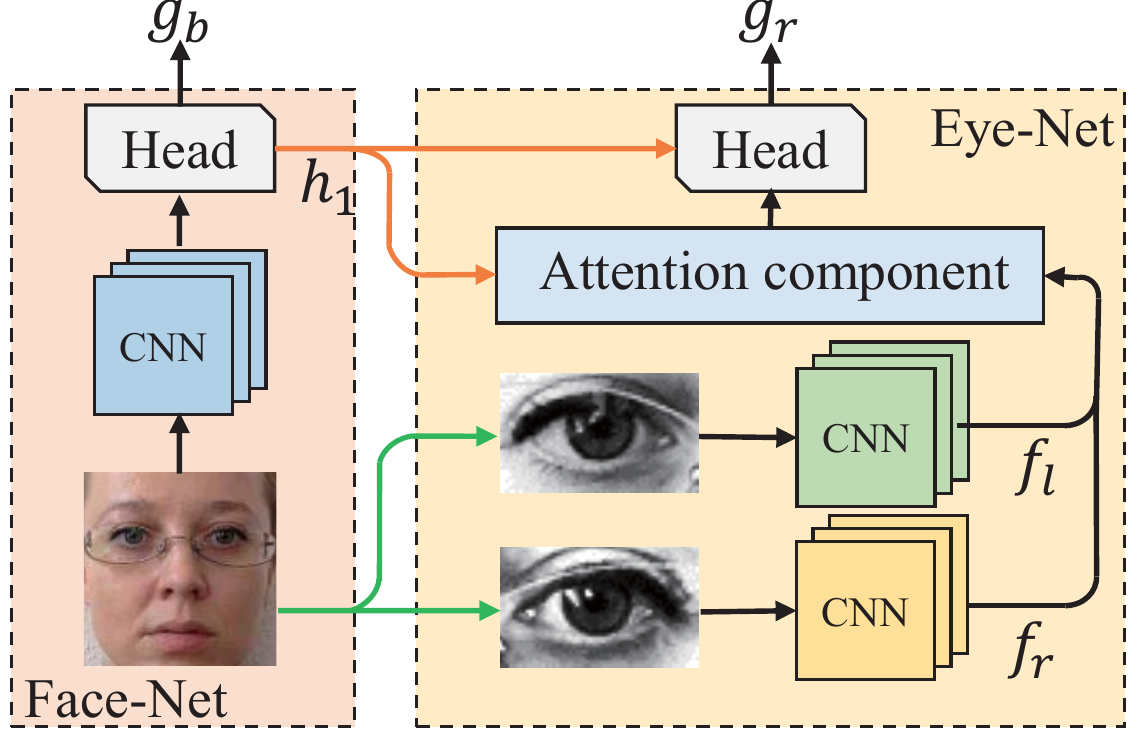}	
	\end{center}
	\caption{The architecture of CA-Net, which estimates gaze in coarse-to-fine way. The Face-Net estimates basic gaze directions from face images. The Eye-Net estimates gaze residuals from eye images. }
	\label{fig:CANet}
\end{figure} 

\subsection{CA-Net}
By integrating above algorithms, we propose CA-Net which can adaptively acquire suitable eye features and estimates 3D gaze directions in a coarse-to-fine way.
The architecture of the proposed CA-Net is shown in~\fref{fig:CANet}.
It contains two subnets, which are Face-Net and Eye-Net.

Face-Net uses face images as input to estimate the basic gaze directions.
We first design a CNN to extract face feature from face images.
After acquiring the face feature, we deliver the face feature into the head component (detail in \fref{fig:coarse2fine}).
The basic gaze direction $g_b$ and state $h_1$ are produced by the head component.

Eye-Net uses two eye images as inputs. 
Two CNNs are designed to extract left eye feature $f_l$ and right eye feature $f_r$.
Then, a attention component is used to fusion $f_l$ and $f_r$ (detail in \fref{fig:attention}). We input the state $h_1$ rather than face feature into the attention component to guiding the generation of eye feature.
After acquiring eye feature, we send the eye feature with $h_1$ into a head component to estimate gaze residuals $g_r$.

The final output of our CA-Net is
\begin{equation}
	\label{equ:output}
	g = g_{b} + g_{r}
\end{equation}

Given the ground truth $g^{*}$, the loss function of CA-Net is defined as 
\begin{equation}
	\label{equ:loss}
	Loss = \alpha*\mathcal{L}(g_{b}, g^{*}) + \beta*\mathcal{L}(g, g^{*}),
\end{equation}
where $\mathcal{L}$ is defined as
\begin{equation}
	\label{equ:angular}
	\mathcal{L}(a, b) = \arccos\left(\frac{a \cdot b}{\|a\| \|b\|}\right).
\end{equation}

We empirically set $\alpha=1$ and $\beta=2$. On the one hand, this loss function encourages CA-Net to estimate an accurate basic gaze direction. On the other hand, we assign a larger weight for $g$ than $g_{b}$ to ensure CA-Net can get a more accurate gaze direction than the basic gaze direction.

\subsection{Implementation detail}

The inputs of CA-Net are 224*224*3 face images, 36*60 gray-scale left and right eye images.

The CNN in Face-Net consists of thirteen convolutional blocks. Each block contains one convolutional layer, one ReLU and one Batch Normalization~\cite{2015_Ioffe_batch}.
The sizes and strides of all convolutional kernels are set as 3*3 and 1. 
The numbers of convolutional kernel are (64, 64, 128, 128, 256, 256, 256, 256, 256, 256, 512, 512, 1024).
We also insert one max pool layer after the second, fourth, seventh and tenth convolutional blocks.
The sizes of max-pooling layers are 2*2 and strides are 2*2.
A global average pooling layer (GAP) is used after the thirteenth block and output 1024D feature.
Final, the 1024D feature is sent to a fully connected layer (FC) to output  256D face features.

The CNN for the left eye in Eye-Net consists of ten convolutional blocks.
The numbers of convolutional kernel are (64, 64 ,128, 128, 128, 256, 256, 256, 512, 1024).
The strides of the second, fifth, eighth convolutional kernels are set as 1.
The sizes of all convolutional kernels are set as 3*3.
Same as Face-Net, a GAP and FC are final used to output the 256D left eye feature.
Meanwhile, for the right eye, the same CNN is designed and output 256D right eye feature.

We implement CA-Net by using Pytorch. We train the whole network in 200 epochs with 32 batch size. The Learning rate is set as 0.001. We initialize the weights of all layers with MSRA initialization~\cite{2015_he_msra}.

\section{Experiment}
\begin{figure}[t]
	\begin{center}
		\includegraphics[width=0.99\columnwidth]{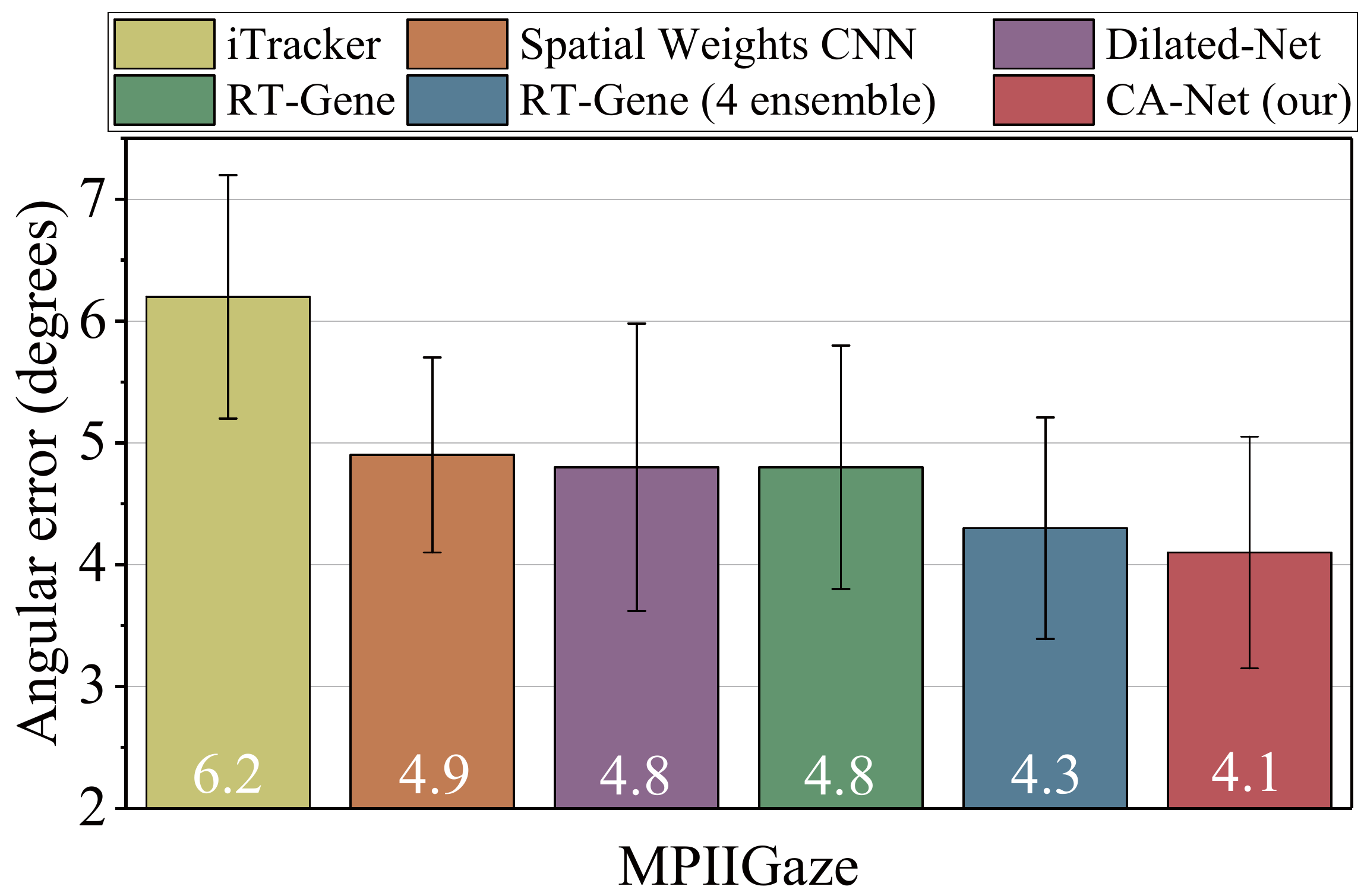}	
	\end{center}
	\caption{Performance in MPIIGaze dataset}
	\label{fig:performance-mpii}
\end{figure}
\subsection{Dataset}
The experiments are conducted in two popular gaze estimation datasets: MPIIGaze~\cite{2017_zhang_gazenet} and EyeDiap~\cite{2014_mora_eyediap}.

MPIIGaze is the largest dataset for appearance-based gaze estimation which provides 3D gaze directions.
It is common used in the evaluation of appearance-based methods~\cite{2017_zhang_fullface,2018_ranjan_light,2018_liu_differential,2018_cheng_asymmetric,2019_xiong_mixed}.
MPIIGaze dataset contains 213,659 images which are captured from 15 subjects.
Note that, MPIIGaze provides a standard evaluation protocol, which selects 3000 images for each subject to compose the evaluation set.
We conduct experiments in the evaluation set rather than the full set. 

EyeDiap dataset contains a set of video clips of 16 participants.
The videos are collected under two visual target sessions, which are screen target and 3D floating ball.
We use the videos collected under screen target sessions and sample one image per fifteen frames to construct the evaluation set.
Note that, since two subjects lack the videos in the screen target session, we obtain the images of 14 subjects finally.

\begin{figure}[t]
	\begin{center}
		\includegraphics[width=\columnwidth]{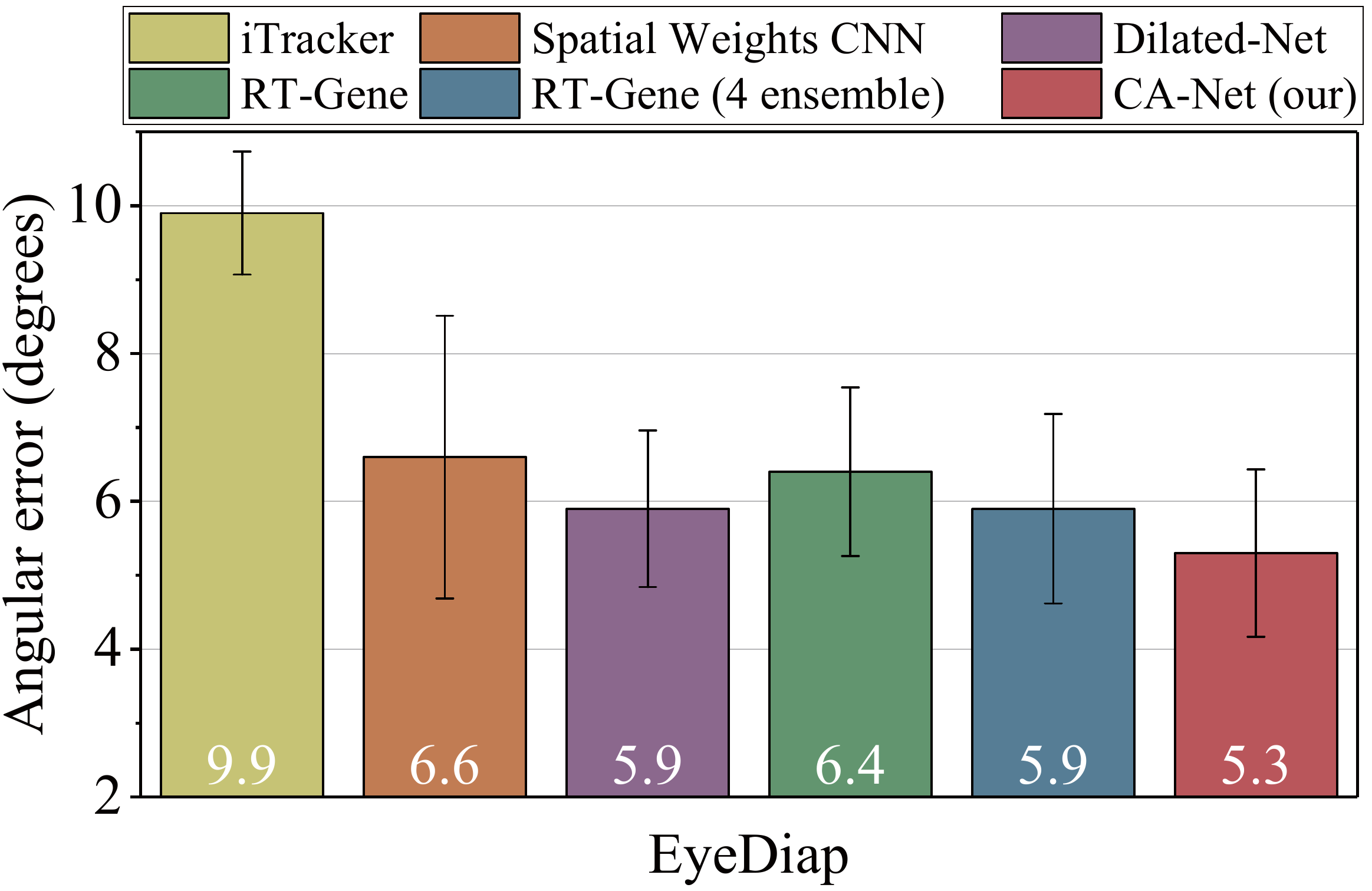}	
	\end{center}
	\caption{Performance in EyeDiap dataset}
	\label{fig:performance-diap}
\end{figure}

\subsection{Data preprocessing}
We follow the process proposed in ~\cite{2017_zhang_gazenet} to normalize the two datasets.
Specifically, the goal of appearance-based gaze estimation is to estimate gaze directions from eye appearances. 
However, since head pose has six freedoms, eye appearances are various in the real world. This complicates the gaze estimation task.
Therefore, we eliminate the translation in head pose by rotating the virtual camera and the roll in head pose by wrapping images.
In addition, we crop eye images from normalized face images with provided landmarks by the dataset.
Note that the landmark can be also automatically detected by various face detection algorithms~\cite{2016_amos_openface}.
The eye images are histogram-equalized and converted into gray scale to eliminate the influence of illumination. 
Note that, the images provided by MPIIGaze has been normalized, we only apply the normalization into EyeDiap.

\subsection{Comparison with appearance based methods}
We first conduct an experiment to compare the performance of the proposed method with other appearance-based methods.
The experiment is conducted in both MPIIGaze and EyeDiap.
Note that, for the two datasets, we both apply the leave-one-person-out strategy to obtain robust results. 

We choose four methods as compared methods, which are \emph{iTracker}~\cite{2016_krafka_itrack}, \emph{Spatial weights CNN}~\cite{2017_zhang_fullface}, \emph{Dilated-Net}~\cite{2018_zhao_dilated} and \emph{RT-Gene}~\cite{2018_tobias_rt}.
Since the accuracy of \emph{RT-Gene} can be improved by four models ensemble, we also show the result of the model ensemble and call it as \emph{RT-Gene (4 ensemble)} to distinguish from \emph{RT-Gene}.
Note that, currently, the best reported performance in MPIIGaze is achieved by \emph{RT-Gene (4 ensemble)}.

\fref{fig:performance-mpii} shows the result in MPIIGaze dataset.
The performances of \emph{Spatial weights CNN}, \emph{Dilated-Net} and \emph{RT-Gene} are all around $4.8^{\circ}$.
The \emph{RT-Gene (4 ensemble)} can improve the performance by a large margin using ensemble scheme, which is $4.3^{\circ}$.    
Our CA-Net achieves the state-of-the-art performance as $4.1^{\circ}$ in the MPIIGaze dataset.
The CA-Net has $0.7^{\circ}$ improvement compared with \emph{RT-Gene} and also has $0.2^{\circ}$ improvement compared with \emph{RT-Gene (4 ensemble)}.
Note that, our CA-Net achieves the state-of-the-art performance without ensemble scheme. The accuracy also can be further improved using ensemble. 

we re-implement the Dilated-Net according to the original paper and use the author provided source codes for the rest methods.
\fref{fig:performance-diap} shows all the results.
\emph{iTracker} has the worst performance because of the shallow network.
\emph{Spatial weights CNN} and \emph{RT-Gene} have $6.6^{\circ}$ and $6.4^{\circ}$ performance in EyeDiap.
However, \emph{Dilated-Net} significantly outperforms \emph{Spatial weights CNN} and \emph{RT-Gene}. It shows the a performance as \emph{RT-Gene (4 ensemble)} which is $5.9^{\circ}$.
Our CA-Net achieves the best performance as $5.3^{\circ}$ in EyeDiap and has $0.6^{\circ}$ improvement compared with \emph{Dilated-Net}. 

The good performance in two datasets demonstrates the advantage of the proposed CA-Net.
In addition, since some recent appearance-based methods do not provide source code and the methods also are difficult to re-implement , we carefully show the reported accuracy in~\Tref{table:performance} for reference.
In order to get a fair comparison, we only show the accuracy in MPIIGaze because the MPIIGaze dataset provides a standard evaluation set. 

\begin{table}[t]
	\renewcommand\arraystretch{1.3}
	\normalsize
	\caption{Comparison between appearance-based methods.}
	\begin{tabular}{c|p{1.5cm}<{\centering}|p{1.5cm}<{\centering}}
		\hline
		Methods& MPIIGaze& 	EyeDiap \\
		\hline
		%\hline
		iTracker									&$6.2^{\circ}$ 		&$9.9^{\circ}$	 	\\
		Spatial Weights CNN									&$4.9^{\circ}$			&$6.6^{\circ}$			\\
		RT-Gene										&$4.8^{\circ}$		&$6.4^{\circ}$ \\
		Dilated-Net									&$4.8^{\circ}$	&$5.9{\circ}$	\\
		RT-Gene(4 ensemble)		 					&$4.3^{\circ}$			&$5.9^{\circ}$			\\
		\hline
		\hline
		Faze~\cite{2019_park_fewshot}				&$5.2^{\circ}$			&$-$ \\
		MeNet~\cite{2019_xiong_mixed}				&$4.9^{\circ}$			&$-$\\
		\hline
		\hline
		Our method									&$\bm{4.1^{\circ}}$	&$\bm{5.3^{\circ}}$	\\
		\hline
	\end{tabular}
	\label{table:performance}
\end{table}

\subsection{Ablation study}
In order to demonstrate the effectiveness of each component in the CA-Net, we perform ablation study in MPIIGaze. 

\textbf{Ablation study about components}.
We first perform ablation study to demonstrate the effect of the attention component and gate component.Specifically, we evaluate two extra methods which are \emph{Gate ablation} and \emph{Attention ablation}.

\emph{Gate ablation} ablates the learned gate from CA-Net, it directly concatenates face feature with eye feature to estimate gaze residuals.
Note that, we do not modify the attention component, where the face feature is also inputted into the attention component to guide the generation of eye feature.

\emph{Attention ablation} ablates the attention component from CA-Net, it assigns fixed weights as $0.5$ for both left eye and right eye to generate the fine-grained eye feature.

The result is shown in the second row of ~\Tref{table:ablation}.
The performance of \emph{Gate ablation} shows $0.32^{\circ}$ decrease compared with CA-Net. 
Meanwhile, the performances of \emph{Attention ablation} have $0.46^{\circ}$ decrease than CA-Net.
It demonstrates the advantages of attention component and learned gate. 

\textbf{Ablation study about network.}
The proposed CA-Net shows the best performance in two datasets. 
However, it is still uncertain whether the coarse-to-fine strategy can improve the performance. 
In order to prove the advantages of coarse-to-fine, we perform ablation study on the network.

We respectively evaluate each subnet in CA-Net. Totally three methods are evaluated. 
\emph{Face-Net.} We directly use the Face-Net to estimate gaze.
\emph{Eye-Net.} We directly use the Eye-Net to estimate gaze from two eye images. Note that, the attention component is not used in this method. We generate eye feature by directly concatenating the left eye feature and right eye feature.
\emph{Joint-Net}.We use the same architecture as CA-Net to extract face feature, left eye feature and right eye feature. The gaze directions are estimated by the joint feature which is generated by concatenating the three features.
We also provide the performance of basic gaze directions in CA-Net and call it as \emph{Face-Net (CA)}.

\begin{table}[t]
	\renewcommand\arraystretch{1.3}
	\normalsize
	\caption{Ablation study.}
	\begin{center}
		\begin{tabular}{|p{4.0cm}<{\centering}|p{3.0cm}<{\centering}|}
			\hline
			Methods & Performance\\
			\hline
			Gate ablation 					&$4.46^{\circ}$				\\
			Attention ablation 			&$4.50^{\circ}$				\\
			
			\hline
			%\hline
			Face-Net				&$4.58^{\circ}$					\\
			Eye-Net				&$5.01^{\circ}$					\\
			Joint-Net		&$5.00^{\circ}$					\\
			Face-Net (CA)	&$4.65^{\circ}$	\\
			\hline
			CA-Net					&$\bm{4.14^{\circ}}$		\\
			\hline
		\end{tabular}
	\end{center}
	\label{table:ablation}
\end{table}

The result is shown in  ~\Tref{table:ablation}.
\emph{Face-Net} shows the best performance between compared methods, which is $4.58^{\circ}$. 
\emph{Face-Net (CA)} achieves $4.65^{\circ}$ performance which has $0.7^{\circ}$ decrease compared with \emph{Face-Net}. However, with the coarse-to-fine strategy, CA-Net achieves $0.51^{\circ}$ improvement than \emph{Face-Net (CA)} and significantly outperforms other methods with $4.14^{\circ}$ performance. This demonstrates the advantages of the proposed coarse-to-fine strategy. 

Moreover, although the backbone of \emph{Joint-Net} is the same as CA-Net, CA-Net achieves $0.86^{\circ}$ improvement than \emph{Joint-Net}.
It is benefited from the proposed coarse-to-fine strategy.

\subsection{Additional analysis}
In order to show the advantages of the algorithms proposed in CA-Net, We perform some additional analysis in MPIIGaze and summarize the results into~\Tref{table:attention}.
The performance of each method is shown in the column of "Refine".
In addition, we also show the performance of basic gaze directions in~\Tref{table:attention} and list the results in the column of "Basic".
We call it as basic performance in the rest parts.

\textbf{Coarse-to-fine v.s. Fine-to-coarse.}
The core of our paper is the coarse-to-fine strategy.
In order to further validate the correctness of the coarse-to-fine strategy, we evaluate the performance of \emph{Fine-to-coarse} .
As for \emph{Fine-to-coarse}, it means to estimate a basic gaze direction from eye images and refine it with residual predicted from face image.

As shown in~\Tref{table:attention}, it is obvious that our CA-Net,~\ie~coarse-to-fine strategy, achieves better performance than \emph{Fine-to-coarse}.
With only changing the strategy, \emph{Fine-to-coarse} has $0.49^{\circ}$ decrease compared with CA-Net.
It demonstrates the advantages of our coarse-to-fine strategy.
In addition, an interesting observation is that the performance of \emph{Fine-to-coarse} is similar with the performance of \emph{Eye-Net} (show in \Tref{table:ablation}) while our CA-Net can improve the performance by a large margin compared with \emph{Face-Net} (show in \Tref{table:ablation}).
It proves the correctness of the proposed coarse-to-fine strategy.

\begin{table}[t]
	\renewcommand\arraystretch{1.3}
	\normalsize
	\caption{Additional analysis about different algorithms.}
	\begin{center}
		
		\begin{tabular}{|p{3.0cm}<{\centering}|p{1.9cm}<{\centering}|p{1.9cm}<{\centering}|}
			\hline
			
			Methods& Basic& 	Refined \\
			\hline
			Fine-to-coarse			&$5.14^{\circ}$		&$5.00^{\circ}$		\\
			
			One gram				&$4.42^{\circ}$ 	&$4.43^{\circ}$	 \\
			
			Face attention				&$4.47^{\circ}$		&$4.30^{\circ}$				\\
			
			Eye attention				&$4.43^{\circ}$		&$4.34^{\circ}$			\\
			\hline
			CA-Net					&$\bm{4.65^{\circ}}$		&$\bm{4.14^{\circ}}$			\\
			\hline
		\end{tabular}
		\label{table:attention}
	\end{center}
\end{table}
\textbf{bi-gram v.s. One gram.}
In order to estimate the gaze direction in a coarse-to-fine way, one key idea is that gaze residuals are associated with basic gaze directions.
Based on the idea, we generalize the coarse-to-fine way as a bi-gram model.
However, it is uncertain whether the bi-gram model is useful.
In this part, we provide a comparison between bi-gram model and one gram model to show the advantages of bi-gram model. 
In particular, we simply use a zero matrix to replace the delivered face feature, where the gaze residuals are only estimated from eye feature.
Note that, we do not modify the attention component. The fine-grained eye feature is also generated with the guiding of face feature.

As shown in ~\Tref{table:attention}, \emph{One gram} shows a better basic performance than CA-Net.
However, without the information about basic gaze directions, the fine-grained eye feature can not further refine the basic gaze direction.  
Final, the \emph{One gram} has $0.29^{\circ}$ decrease than CA-Net.
The result demonstrates the usefulness of bi-gram model.

\textbf{Attention component v.s. other weight generations}
In order to acquire suitable fine-grained feature to estimate gaze residuals,
we propose an attention component to adaptively assign weights for left and right eyes.
Specifically, the attention component learns the eye weights from face feature and corresponding eye feature.
In order to show the advantages of the proposed attention component, in this part, we conduct comparison by replacing the weight component.  

There are two weight generations chosen for comparing.
\emph{Face attention} generates the weights of two eyes from face feature.
\emph{Eye attention} generates the weights of two eyes from corresponding eye features.
The results are shown in~\Tref{table:attention}.
A suitable baseline is \emph{Attention ablation} (show in \Tref{table:ablation}), which achieve $4.5^{\circ}$ performance.
As shown in ~\Tref{table:attention}, \emph{Face attention} and \emph{Eye attention} show the better performance compared with \emph{Abalte Attention}.
It demonstrates that the face feature and corresponding eye feature both are useful for the coarse-to-fine gaze estimation.
Meanwhile, they both show worse performance than CA-Net.
This demonstrates the advantages of the proposed attention component. 

\begin{figure}[t]
	\begin{center}
		\includegraphics[width=0.99\columnwidth]{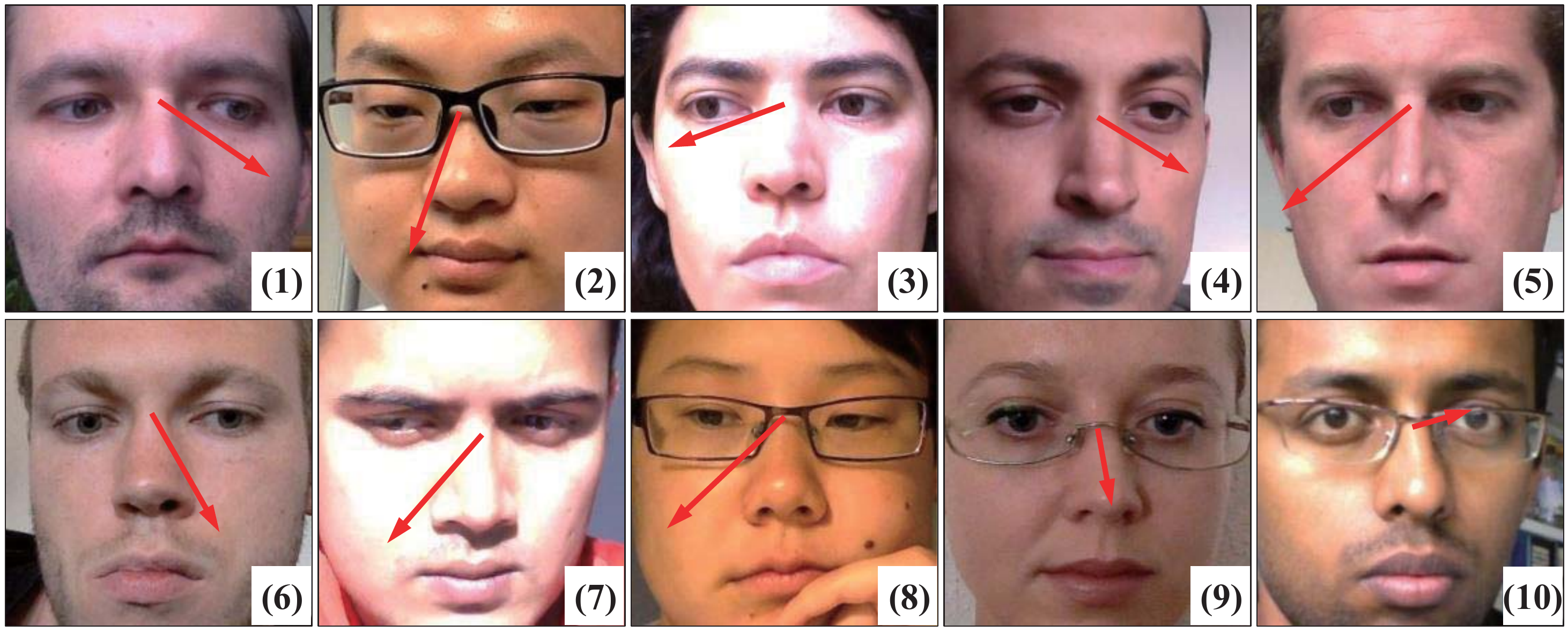}	
	\end{center}
	\caption{Some visual results of estimated 3D gaze.}
	\label{fig:casestudy}
\end{figure}

\textbf{Visual results.}
We also show some visual results in \fref{fig:casestudy}.
It is obvious that our method can perform well in different cases.
In addition, as shown in the sixth and seventh sub-figures in \fref{fig:casestudy}, our CA-Net can also produce accurate gaze directions when the gaze direction deviates from the face direction.
This demonstrates that our method not only focuses on face images but also is sensitive to the eye region.

\section{Conclusion}
In this paper, we propose a coarse-to-fine strategy to estimate gaze directions.
The process of the coarse-to-fine strategy is to estimate a basic gaze direction from face image and refine it with residual predicted from eye images.
A key point of the coarse-to-fine strategy is the estimation of gaze residuals.
In order to accurately estimate gaze residuals, we propose an attention component to adaptively assign weights for eye images and to obtain suitable eye feature.
In addition, we also generalize the coarse-to-fine process as a bi-gram model to bridge the basic gaze directions and gaze residuals. 
Based on above algorithms, we propose CA-Net, which can adaptively acquire suitable fine-grained feature and estimates 3D gaze directions in a coarse-to-fine way.
Experiments show the CA-Net achieves state-of-the-art performance in MPIIGaze and EyeDiap.
\bibliographystyle{aaai}
\bibliography{egbib}

\end{document}